\documentclass{article}

\usepackage[square,sort,comma,numbers]{natbib}

\usepackage[preprint]{nips_2018}

\usepackage[utf8]{inputenc} % allow utf-8 input
\usepackage[T1]{fontenc}    % use 8-bit T1 fonts
\usepackage{hyperref}       % hyperlinks
\usepackage{url}            % simple URL typesetting
\usepackage{booktabs}       % professional-quality tables
\usepackage{amsfonts}       % blackboard math symbols
\usepackage{nicefrac}       % compact symbols for 1/2, etc.
\usepackage{microtype}      % microtypography

\usepackage{multirow}
\usepackage{color,soul}
\usepackage{mathtools}
\usepackage{graphicx}

\newcommand{\zmat}[1]{\MakeUppercase{#1}}
\newcommand{\zvec}[1]{\boldsymbol{#1}}
\newcommand{\zset}[1]{\big\{ #1 \big\}}

\usepackage{epstopdf}

\title{SRP: Efficient class-aware embedding learning for large-scale data via supervised random projections}

\author{
  Amir-Hossein Karimi$^{1}$,
  Alexander Wong$^2$,
  Ali Ghodsi$^3$
  \\
  $^1$David Cheriton School of Computer Science,  University of Waterloo \\
  $^2$Systems Design Engineering, University of Waterloo \\
  $^3$Department of Statistics and Actuarial Science, University of Waterloo \\
  \texttt{\{a6karimi, a28wong, aghodsib\}@uwaterloo.ca} \\
}

\begin{document}

\maketitle

\begin{abstract}

  % Supervised dimensionality reduction strategies have been of great interest.  However, current supervised dimensionality reduction approaches are difficult to scale for situations characterized by large datasets given the high computational complexities associated with a core step: spectral decomposition.  In this study we explore a novel direction of directly learning optimal class-aware embeddings in a supervised manner via the notion of supervised random projections (SRP), inspired by Supervised Principal Component Analysis (SPCA) and by leveraging randomized Kernel Approximations.  Experimental results on five synthetic and real-world datasets demonstrate that the proposed SRP strategy for class-aware embedding learning can be very promising in producing embeddings that are highly competitive with existing supervised dimensionality reduction methods (e.g., SPCA and KSPCA) while achieving 1-2 orders of magnitude better computational performance.

  Supervised dimensionality reduction strategies have been of great interest.  However, current supervised dimensionality reduction approaches are difficult to scale for situations characterized by large datasets given the high computational complexities associated with such methods.  While stochastic approximation strategies have been explored for unsupervised dimensionality reduction to tackle this challenge, such approaches are not well-suited for accelerating computational speed for supervised dimensionality reduction.  Motivated to tackle this challenge, in this study we explore a novel direction of directly learning optimal class-aware embeddings in a supervised manner via the notion of supervised random projections (SRP).  The key idea behind SRP is that, rather than performing spectral decomposition (or approximations thereof) which are computationally prohibitive for large-scale data, we instead perform a direct decomposition by leveraging kernel approximation theory and the symmetry of the Hilbert-Schmidt Independence Criterion (HSIC) measure of dependence between the embedded data and the labels.  Experimental results on five different synthetic and real-world datasets demonstrate that the proposed SRP strategy for class-aware embedding learning can be very promising in producing embeddings that are highly competitive with existing supervised dimensionality reduction methods (e.g., SPCA and KSPCA) while achieving 1-2 orders of magnitude better computational performance.  As such, such an efficient approach to learning embeddings for dimensionality reduction can be a powerful tool for large-scale data analysis and visualization.

\end{abstract}

\section{Introduction}
\label{introduction}

Consider the supervised task of predicting a dependent response random variable for an independent high-dimensional explanatory random variable. Conventional classification and regression methods are susceptible to the ``curse of dimensionality'' where satisfactory results depend on high data dimensionality which in turn requires an exponentially large number of data points. To combat this curse for a dataset, $\zmat{X} = [\zvec{x}_1, \zvec{x}_2, \cdots, \zvec{x}_n] \in \mathbb{R}^{d \times n}$, it is often desirable to find a low-dimensional representation of the dataset to be used for downstream processing. Conventional dimensionality reduction methods such as Principal Component Anaylsis (PCA) are unsupervised and result in an embedding that preserves directions of maximum variation in the data. However, in many cases, the interesting directions of variation align with the labels, $\zmat{Y} = [\zvec{y}_1, \zvec{y}_2, \cdots, \zvec{y}_n]\in \mathbb{R}^{\ell \times n}$, that accompany the data.

Many methods exist for supervised dimensionality reduction that guide the algorithm toward the modes of variability that are of particular interest. These methods include Fisher's Discriminant Analysis FDA \cite{fisher1936use}, the large family of methods known as Metric Learning \cite{xing2003distance} \cite{weinberger2006distance}, the family of Sufficient Dimensionality Reduction (SDR) algorithms \cite{fukumizu2004dimensionality} \cite{li1991sliced}, and Supervised Principal Components proposed by Bair et. al. \cite{bair2006prediction}. Another approach is Supervised Principal Component Analysis (SPCA) \cite{barshan2011supervised}, a generalization of PCA which finds a linear embedding of the data, $\zmat{U}^T \zmat{X}$, that has maximum dependence on the labels, $\zmat{Y}$. This method has demonstrated superior performance compared to other methods on tasks ranging from regression to classification to visualization, and benefits from having a closed-form solution that can be obtained by Singular Value Decomposition (SVD).

The reliance on SVD, however, limits the applicability of SPCA and others on large datasets. Specifically, the SVD of a dense matrix, $\zmat{X} \in \mathbb{R}^{d \times n}$, requires $\mathcal{O}(\max\{dn^2, d^2n\})$ operations for full decomposition \cite{stewart2000decompositional}. To only obtain the eigenvectors associated with the top-$k$ eigenvalues, this time is reduced to $\mathcal{O}(dnk)$ \cite{holmes2007fast}. Even more efficient approaches that based on randomized algorithms provide approximate solutions in $\mathcal{O}(dn\log(k))$ \cite{halko2011finding} with tight guarantees, which is still burdensome for large-scale datasets. We present a solution that bypasses SVD altogether.

In the present work, we set out the design a randomized approximate method for performing label-aware Supervised Principal Component Analysis. Whereas there is a vast literature on randomized approaches to PCA (see \cite{halko2011finding} and \cite{rokhlin2009randomized}), we are not aware of any such approach for SPCA. Furthermore, in our experiments we found that while randomized SVD improved on the run-time of SPCA's decomposition stage, it had a relatively small effect on the overall time performance when considering all stages of SPCA. In this work, we propose a completely new approach to directly obtain the optimal subspace that has maximum dependence between the embedded data and the labels. This presented approach is based on the principals of kernel approximation, and therefore, in Section \ref{methodology} we review the fundamentals of SPCA and kernel approximation, upon which the proposed work is based. Here we also provide a detailed analysis of time complexity and provide convergance bounds of guarantees for the proposed method. In Section \ref{experiments}, we detail the experimental setup and discuss the results. Finally, in Section \ref{conclusion} conclude with suggestions for future steps.

\section{Methodology}
\label{methodology}

In this section, we will review the foundations of Supervised Principal Component Analysis (SPCA), and briefly review the literature on kernel approximations. We show how the combination of these approaches leads to a novel approach for performing supervised class-aware embeddings very efficiently without the need for Singular Value Decomposition.

\subsection{Supervised Principal Component Analysis}
\label{supervised_principal_component_analysis}

Suppose we have a dataset $\mathcal{S} = \zset{(\zvec{x_i}, \zvec{y_i})}_{i=1}^n \subseteq \mathcal{X} \times \mathcal{Y}$ where $\mathcal{X} \in \mathbb{R}^d$ is the space of all $d$-dimensional explanatory variables, $\mathcal{Y} \in \mathbb{R}^\ell$ is the space of all $\ell$-dimensional response variables. Let $\zmat{X} \in \mathbb{R}^{d \times n}$ and $\zmat{Y} \in \mathbb{R}^{\ell \times n}$ be particular realizations of $n$ random pairs of variables sampled independently from $P_{\mathcal{X}, \mathcal{Y}}$. We aim to find an orthogonal projection $\zmat{U}$ of $\zmat{X}$ to maximize the dependence of $\zmat{U}^T\zmat{X}$ on $\zmat{Y}$.

To maximize the dependence of $\zmat{Y}$ on $\zmat{U}^T\zmat{X}$, we must consider all forms of dependence, including linear and nonlinear variants. It is commonly known that linear dependence between a pair of random variables can be measured as the cross-correlation between those random variables. On the other hand, nonlinear dependence can be captured by looking at the cross-correlation between all nonlinear transformations of those random variables. Clearly, however, there is a problem with evaluating all nonlinear transformations as there can be infinitely many of them. To overcome this, we turn to the commonly used Hilbert-Schmidt Independence Criterion (HSIC) introduced by Gretton et al. \cite{gretton2005measuring}. HSIC essentially projects the random variables from an original space into an abstract Reproducing Kernel Hilbert Space (RKHS), defines cross-correlation in that space, and evaluates dependence in a closed-form manner.
Therefore, HSIC is an effective tool for ``measuring'' (linear and nonlinear) dependence between two random variables. While the exact value of HSIC is measured by computing the cross-covariance between $\mathcal{F}$ and $\mathcal{G}$ (where $\mathcal{F}$ and $\mathcal{G}$ are separable RKHS containing all continuous bounded real-valued functions of $x$ from $\zmat{X}$ to $\mathbb{R}$ and $y$ from $\zmat{Y}$ to $\mathbb{R}$, respectively), empirical approximations to the HSIC value between random variables $\zmat{X}$ and $\zmat{Y}$ can simply be calculated by evaluating the following on the observations in the population:

\vspace{-5mm}

\begin{align}
  HSIC(\mathcal{S}, \mathcal{F}, \mathcal{G}) = \frac{ \mathbf{tr}(\zmat{K} \zmat{H} \zmat{L} \zmat{H}) }{(n-1)^2}
\end{align}

\vspace{-2mm}

\noindent where $\zmat{H}, \zmat{K}, \zmat{L} \in \mathbb{R}^{n \times n}$, $\zmat{K}_{ij} \coloneqq k(\zvec{x_i}, \zvec{x_j})$, $\zmat{L}_{ij} \coloneqq l(\zvec{y_i}, \zvec{y_j})$, and $\zmat{H}_{ij} \coloneqq \zmat{I} - \zvec{e} \zvec{e}^T / n$ is the centering matrix. Therefore, to maximize the dependence between $U^T\zmat{X}$ and $\zmat{Y}$, $\zmat{K}$ is set to the kernel of $U^T\zmat{X}$ and therefore we must maximize the following:

\vspace{-5mm}

\begin{align}
  \mathbf{tr}(\zmat{K} \zmat{H} \zmat{L} \zmat{H}) &= \mathbf{tr}(\zmat{X}^T \zmat{U} \zmat{U}^T \zmat{X} \zmat{H} \zmat{L} \zmat{H}) \\
                                                   &= \mathbf{tr}(\zmat{U}^T \zmat{X} \zmat{H} \zmat{L} \zmat{H} \zmat{X}^T \zmat{U})
\end{align}

\vspace{-2mm}

\noindent where the second line is via properties of trace. This optimization problem, however, is ill-defined as it is unbounded above. To construct the final optimization problem, we add the commonly used condition for orthogonality of the transformation matrix $\zmat{U}$ (incidentally, this condition makes the optimization problem well-defined by bounding the objective function), and we obtain the following:

\vspace{-3mm}

\begin{equation}
  \label{eq:spca_optimization_problem}
  \begin{aligned}
    \underset{\zmat{U}}{\text{argmax}} & \quad \mathbf{tr}(\zmat{U}^T \zmat{X} \zmat{H} \zmat{L} \zmat{H} \zmat{X}^T \zmat{U}) \\
    \textit{subject to}                & \quad \quad \quad \quad \zmat{U}^T \zmat{U} = \zmat{I}                                \\
  \end{aligned}
\end{equation}

\vspace{-1mm}

\noindent This optimal $\zmat{U}$ that solved Eq. \eqref{eq:spca_optimization_problem} are the eigenvectors corresponding to the top-$k$ eigenvalues of $\zmat{Q} = \zmat{X} \zmat{H} \zmat{L} \zmat{H} \zmat{X}^T$. Because $\zmat{Q}$ is a real, symmetric, and positive semidefinite matrix, the top eigenvectors can be obtained in closed-form via Singular Value Decomposing (SVD).

This approach is called \textit{Supervised Principal Component Analysis} (SPCA) \cite{barshan2011supervised}. Nonlinear extensions of SPCA can be formulated by expressing the transformation matrix $\zmat{U}$ as a linear combination of the projected data points, $\zmat{U} = \Phi(\zmat{X}) \zmat{\beta}$, via representation theory \cite{alperin1993local}. Plugging $\zmat{U}$ back into Eq. \eqref{eq:spca_optimization_problem}, we obtain a new optimization problem:

\vspace{-5mm}

\begin{equation}
  \label{kspca_optimization_problem}
  \begin{aligned}
    \underset{\zmat{U}}{\text{argmax}} & \quad \mathbf{tr}(\zmat{\beta}^T \zmat{K} \zmat{H} \zmat{L} \zmat{H} \zmat{K}^T \zmat{\beta}) \\
    \textit{subject to}                & \quad \quad \quad \quad \zmat{\beta}^T \zmat{K} \zmat{\beta} = \zmat{I}                       \\
  \end{aligned}
\end{equation}

\noindent where $\zmat{K} = \Phi(\zmat{X})^T \Phi(\zmat{X})$ is the kernel matrix of the data, $\zmat{X}$. The solution, $\zmat{\beta}$, for \textit{Kernel Supervised Principal Component Analysis} (KSPCA) can be ontained by solving the generalized eigenvector problem above, and obtained by decomposing $\zmat{Q} = \zmat{H} \zmat{L} \zmat{H} \zmat{K}$ via SVD.

Now that we have established the derivation of SPCA and KSPCA, it is worth restating that computing the eigenvectors corresponding to the top-$k$ eigenvalues of $\zmat{Q}$ ($\in \mathbb{R}^{d \times d}$ for SPCA, and $\in \mathbb{R}^{n \times n}$ for KSPCA) is computationally burdensome for large datasets. See Table \ref{table:time_complexities} for a detailed time analysis. In the next section, we propose an alternative approach to solving Eq. \ref{eq:spca_optimization_problem} that allows us to bypass SVD completely. For this, we focus specifically on the symmetrical form of $\zmat{Q} = \zmat{X} \zmat{H} \zmat{L} \zmat{H} \zmat{X}^T$.

\subsection{Supervised Random Projections}
\label{supervised_random_projections}

\noindent \textbf{Claim 1} \textit{Let $\zmat{Z}_1 = \zmat{U}^T \zmat{X}$ be the embedding obtained by SPCA and $\zmat{Z}_2 = \zmat{\Psi} \zmat{H} \zmat{X}^T \zmat{X}$, where $\zmat{X} \in \mathbb{R}^{d \times n}$ is the data matrix, $\zmat{H} \in \mathbb{R}^{n \times n}$ is the centering matrix, and $\zmat{\Psi}  \in \mathbb{R}^{k \times n}$ is a decomposition of the positive semidefinite matrix such that $\zmat{L} = \zmat{\Psi}^T \zmat{\Psi}$. Suppose further that $\zmat{U}$ and $\zmat{\Sigma}$ are the eigenvectors and eigenvalues of $\zmat{Q} = \zmat{X} \zmat{H} \zmat{L} \zmat{H} \zmat{X}^T$, respectively. It can be shown that $\zmat{Z}_2 = \zmat{\Sigma}^\frac12 \zmat{Z}_1$ up to a rotation.}

% \noindent \textbf{Proof:} Starting with the SVD of real, symmetric, and positive semidefinite matrix, $\zmat{Q}$, we have:

% \vspace{-5mm}

% \begin{align}
%   \zmat{Q} &= \zmat{U} \zmat{\Sigma} \zmat{U}^T \\
%            &= \zmat{U} \zmat{\Sigma}^\frac12 \zmat{R}^T \zmat{R} \zmat{\Sigma}^\frac12 \zmat{U}^T
% \end{align}

% \vspace{-2mm}

% \noindent where $\zmat{R}$ is an orthonormal rotation matrix. On the other hand, we have:

% \vspace{-5mm}

% \begin{align}
%   \zmat{Q} &= \zmat{X} \zmat{H} \zmat{L} \zmat{H} \zmat{X}^T \\
%            &= \zmat{X} \zmat{H} \zmat{\Psi}^T \zmat{\Psi} \zmat{H} \zmat{X}^T
% \end{align}

% \vspace{-2mm}

% Therefore, we can conclude $\zmat{R} \zmat{\Sigma}^\frac12 \zmat{U}^T = \zmat{\Psi} \zmat{H} \zmat{X}^T$, and:

\noindent \textbf{Proof:} Starting with the SVD of real, symmetric, and positive semidefinite matrix, $\zmat{Q}$, we have:

\vspace{-5mm}

\noindent\begin{minipage}[t]{.5\linewidth}
  \begin{align}
    \zmat{Q} &= \zmat{U} \zmat{\Sigma} \zmat{U}^T \\
             &= \zmat{U} \zmat{\Sigma}^\frac12 \zmat{R}^T \zmat{R} \zmat{\Sigma}^\frac12 \zmat{U}^T
  \end{align}
\end{minipage}%
\begin{minipage}[t]{.5\linewidth}
  \begin{align}
    \zmat{Q} &= \zmat{X} \zmat{H} \zmat{L} \zmat{H} \zmat{X}^T \\
             &= \zmat{X} \zmat{H} \zmat{\Psi}^T \zmat{\Psi} \zmat{H} \zmat{X}^T
  \end{align}
\end{minipage}

\noindent where $\zmat{R}$ is an orthonormal rotation matrix. Therefore, we can conclude $\zmat{R} \zmat{\Sigma}^\frac12 \zmat{U}^T = \zmat{\Psi} \zmat{H} \zmat{X}^T$, and:

\vspace{-5mm}

\begin{align}
  \implies \zmat{Z}_2 &= \zmat{\Psi} \zmat{H} \zmat{X}^T \zmat{X} \\
                      &= \zmat{R} \zmat{\Sigma}^\frac12 \zmat{U}^T \zmat{X} \\
                      &= \zmat{R} \zmat{\Sigma}^\frac12 \zmat{Z}_1 \quad \square
\end{align}

\vspace{-2mm}

Interestingly, we can obtain a $\zmat{\Psi}$ to approximate $\zmat{L} = \zmat{\Psi}^T \zmat{\Psi}$ simply through the use of kernel approximations (Section \ref{kernel_approximation}). The method presented in this section will henceforth be referred to as \textit{Supervised Random Projections} (SRP). In order to embed the data into $k$ dimensions, SPCA constructed a matrix, $\zmat{U}$, whose columns were the eigenvectors corresponding to the top-$k$ eigenvalues of $\zmat{Q}$. In SRP, however, we have $\zmat{\hat{U}}_{d \times k} = \zmat{X}_{d \times n} ~ \zmat{H}_{n \times n} ~ \zmat{\Psi}^T_{n \times k}$, where $\zmat{\Psi}$ is obtained via a rank-k approximation of $\zmat{L} = \zmat{\Psi}^T \zmat{\Psi}$, and the embedding is obtained via

\vspace{-5mm}

\begin{align}
  \zmat{\hat{X}} &= \zmat{\hat{U}}^T \zmat{X} \\
                 &= \zmat{\Psi} \zmat{H} \zmat{X}^T \zmat{X}
\end{align}

\vspace{-3mm}

\noindent A simple extension of the above linear down-projection is to replace $\zmat{X}^T \zmat{X}$ with a kernelized version of the data, i.e., $\zmat{\hat{X}} = \zmat{\Psi} \zmat{H} \zmat{K}$, which shall be referred to as \textit{Kernel Supervised Random Projections} (KSRP). To obtain rank-k approximations of $\zmat{L}$, we use randomized kernel strategies, detailed next.

It is noteworthy that PCA is itself a special case of SPCA when labels are either not present, or not used \footnotemark. Therefore, the proposed Supervised Random Projections method is also a novel approach for performing PCA in a randomized manner, namely \textit{Randomized Principal Component Analysis}. Essentially for many applications, the proposed method makes PCA tractable on large datasets. We defer these derivations to future work.

\footnotetext{
  \noindent\begin{minipage}[t]{.65\linewidth}
    In such a case, the $\zmat{L}$ kernel is set equal to the identity matrix, i.e., a kernel which only captures the similarity between a point and itself. Therefore, (from \cite{barshan2011supervised}), $\zmat{Q}$ becomes the covariance matrix of mean-subtracted samples $\zmat{X}$, and decomposing the covariance matrix is the same as decomposing $\zmat{Q}$ and consequently the same as maximizing $\mathbf{tr}(\zmat{U}^T \zmat{Q} \zmat{U})$. In other words, setting $\zmat{L} = \zmat{I}$ means that we retain the maximal diversity between observations, and therefore PCA is a special case of SCPA.
  \end{minipage}%
  ~ ~ \begin{minipage}[t]{.3\linewidth}
    \begin{align*}
      \zmat{Q} &= \zmat{X} \zmat{H} \zmat{L} \zmat{H} \zmat{X}^T                                                              \\
               &= \zmat{X} \zmat{H} \zmat{I} \zmat{H} \zmat{X}^T                                                              \\
               &= (\zmat{X} \zmat{H}) (\zmat{X} \zmat{H})^T                                                                   \\
               % &= [\zmat{X} (\zmat{I} - \frac{\zvec{e}\zvec{e}^T}{n})] [\zmat{X} (\zmat{I} - \frac{\zvec{e}\zvec{e}^T}{n})]^T \\
               &= (\zmat{X} - \mu_{x}) (\zmat{X} - \mu_{x})^T                                                                 \\
               &= Cov(\zmat{X})
    \end{align*}
  \end{minipage}
}

\begin{table*}
  \caption[Comparing time complexities of SPCA, KSPCA, SRP, and KSRP]{Comparing time complexities of SPCA, KSPCA, SRP, and KSRP. Reminder: $\zmat{X}^{d \times n}, ~ \zmat{Y}^{\ell \times n}, ~ \zmat{K}^{n \times n}, ~ \zmat{L}^{n \times n}, ~ \zmat{H}^{n \times n}, ~ \zmat{\Psi_{\zmat{X}}}^{k_x \times n}, ~ \zmat{\Psi_{\zmat{Y}}}^{k_y \times n}$ where $n$ is the number of training samples, $d$ is the original data dimensionality, $\ell$ is the label dimensionality, and $k_x$ and $k_y$ are the explicit embedding space dimensionality or the number of random bases used to approximate $\zmat{K} = \zmat{\Psi}_{\zmat{X}}^T \zmat{\Psi}_{\zmat{X}}$ and $\zmat{L} = \zmat{\Psi}_{\zmat{Y}}^T \zmat{\Psi}_{\zmat{Y}}$, respectively. Because $k_y$ determines the dimension of the embedding space (see text), we have $k_y < d$. We set $k_x = 1000$ to well-approximate the data kernel $\zmat{K}$; this value does not effect the dimensionality of the embedding space. Finally, it is assumed we have more data than dimensions: $d < n$.}
  % The subscript under $\zmat{\Psi}$ refers to whether we are approximating the data kernel $\zmat{K}$ (via explicit mapping $\zmat{\Psi}_{\zmat{X}}$ with $k_x$ random bases) or the labels kernel $\zmat{L}$ (via explicit mapping $\zmat{\Psi}_{\zmat{Y}}$ with $k_y$ random bases).}
  \label{table:time_complexities}
  \centering

  \begin{tabular}{c | c | cc | cc} % C{2.25cm} C{2.25cm} C{2.25cm} C{2.25cm}}
    \hline

    % Method                                                                             & Matrix Multiplication                   & Kernel Computation ($\zmat{K},\zmat{\Psi}$) & SVD                \\
    \multirow{2}{*}{Method}                                                            & \multirow{2}{*}{Matrix Mult} & \multicolumn{2}{c|}{Kernel Computation}                                 & \multicolumn{2}{c}{SVD (top-$k$)}                 \\
                                                                                       &                              & ($\zmat{L}, \zmat{\Psi_\zmat{Y}}$) & ($\zmat{K}, \zmat{\Psi_\zmat{X}}$) & exact                & approx                     \\

    \hline

    $\zmat{U}_{SPCA}       = eig(\zmat{X} \zmat{H} \zmat{L} \zmat{H} \zmat{X}^T)$      & $\mathcal{O}(n^3)$           & $\mathcal{O}(\ell n^2)$            & -                                  & $\mathcal{O}(d^2 k)$ & $\mathcal{O}(d^2 \log(k))$ \\
    $\zmat{U}_{KSPCA}      = eig(\zmat{H} \zmat{L} \zmat{H} \zmat{K})$                 & $\mathcal{O}(n^3)$           & $\mathcal{O}(\ell n^2)$            & $\mathcal{O}(d n^2)$               & $\mathcal{O}(n^2 k)$ & $\mathcal{O}(n^2 \log(k))$ \\
    $\zmat{\hat{U}}_{SRP}  = \zmat{X} \zmat{H} \zmat{\Psi}^T_{\zmat{Y}}$               & $\mathcal{O}(d n^2)$         & $\mathcal{O}(k_y d n)$             & -                                  & -                    & -                          \\
    $\zmat{\hat{U}}_{KSRP} = \zmat{\Psi}_{\zmat{X}} \zmat{H} \zmat{\Psi}^T_{\zmat{Y}}$ & $\mathcal{O}(k_x n^2)$       & $\mathcal{O}(k_y d n)$             & $\mathcal{O}(k_x d n)$             & -                    & -                          \\

    \hline
  \end{tabular}
\end{table*}

\subsection{Kernel Approximation}
\label{kernel_approximation}

Kernel methods are successful techniques used broadly in many machine learning problems \cite{scholkopf2002learning}. Despite the success of these methods, kernel methods have limited applicability in large-scale problems due to poor scaling in the face of increasing number of training samples. This problem, commonly known as the \textit{curse of support}, presents itself when storing the Kernel matrix, and more importantly at test time when the Kernel matrix is used to evaluate a decision function for a new test sample. In their seminal work, Rahimi and Recht \cite{rahimi2007random} suggested that by mapping the data (both train and test) into a relatively low-dimensional randomized feature space, one can operate on an explicit lower-dimensional space satisfying:

\vspace{-5mm}

\begin{align}
  k(\zvec{x},\zvec{y}) = \langle \phi(\zvec{x}), \phi(\zvec{y}) \rangle \approx \zvec{\psi}(\zvec{x})^T \zvec{\psi}(\zvec{y})
\end{align}

\noindent where the parameters of $\zvec{\psi}$ are random bases sampled independently from the inverse Fourier transform of the desired shift-invariant kernel (see Bochner's theorem). Incidentally, this covers a wide class of kernel functions including Gaussian RBF, Laplace, Matern, etc. Thus, instead of evaluating the entries of the kernel matrix individually, the entire kernel matrix $\zmat{K}$ can be approximated via a fixed set of random bases drawn from the above distribution applied to the data samples. This method came to be known as \textit{Random Fourier Features} and was later extended and referred to as \textit{Random Kitchen Sinks} \cite{rahimi2007random}.

To summarize, we set out to find $\zmat{U}^T \zmat{X}$, a transformation of the data $\zmat{X}$ that had maximum dependence with the labels $\zmat{Y}$. This problem was formulated as on optimization problem \eqref{eq:spca_optimization_problem}, subject to simple constraints, the solution for which was initially obtained via SVD, but now has a direct formulation using random kernel approximations.

\section{Experiments}
\label{experiments}

In this section we study the effectiveness of the proposed Supervised Random Projections (SRP) and Kernel Supervised Random Projections (KSRP) methods in comparison with Supervised Principal Component Analysis (SPCA) and Kernel Supervised Principal Component Analysis (KSPCA). These methods are compared on a number of visualization and classification problems, assessing their embedding performance using metrics such 1-Nearest Neighbor classification performance (common for evaluating embedding quality; see \cite{van2009dimensionality}), and wall-clock duration measurements.

In all of the following experiments, the input features are first normalized to the range [0, 1]. Wherever a data kernel was used in the methods above, $\zmat{K}$ was an RBF kernel with variance $\sigma_{\zmat{X}}$ obtained using 10-fold cross-validation. For the labels kernel, we apply a delta kernel $\zmat{L}(\zvec{p}, \zvec{q}) = \delta(\zvec{p}, \zvec{q})$ to compute $\zmat{L}$. This choice of kernel results in embeddings where instances of the same class are grouped together, as desired. In order to apply kernel approximation techniques for the delta kernel, we simply approximate $\zmat{L}$ using an RBF kernel with a very small variance. In our experiments, we use $\sigma_{\zmat{Y}} = 10^{-10}$.

A critical element of SRP and KSRP is the number of random bases used to approximate the kernels $\zmat{K} = \zmat{\Psi}_{\zmat{X}}^T \zmat{\Psi}_{\zmat{X}}$ and $\zmat{L} = \zmat{\Psi}_{\zmat{Y}}^T \zmat{\Psi}_{\zmat{Y}}$. Because $k_y$ determines the dimension of the embedding space (see Section \ref{supervised_random_projections}), we set $k_y = k$, the desired dimensionality; therefore, $k_y < d$. We set $k_x = 1000$ to well-approximate the data kernel $\zmat{K}$; this value does not effect the dimensionality of the embedding space. Refer to Table \ref{table:time_complexities} for detailed analysis of the effect of these parameters on time complexity. In fact, for KSRP, we can use the actual kernel $\zmat{K}$ instead of an approximate, but an approximate is more efficient computationally.

\begin{figure*}[t!]
  $\begin{array}{rl}
      \fbox{\includegraphics[height=0.36\textwidth, width=0.45\textwidth]{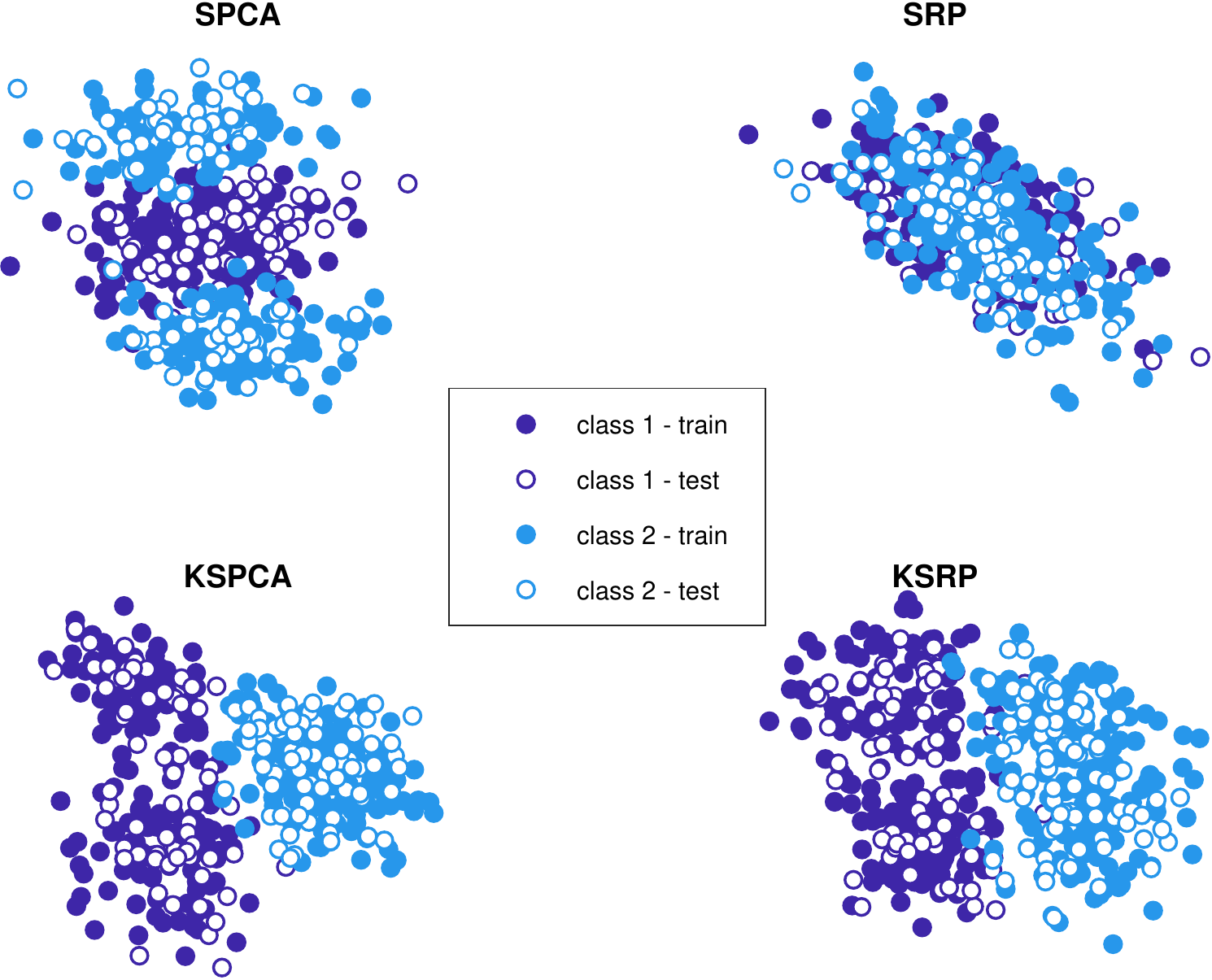}} &
      \fbox{\includegraphics[height=0.36\textwidth, width=0.45\textwidth]{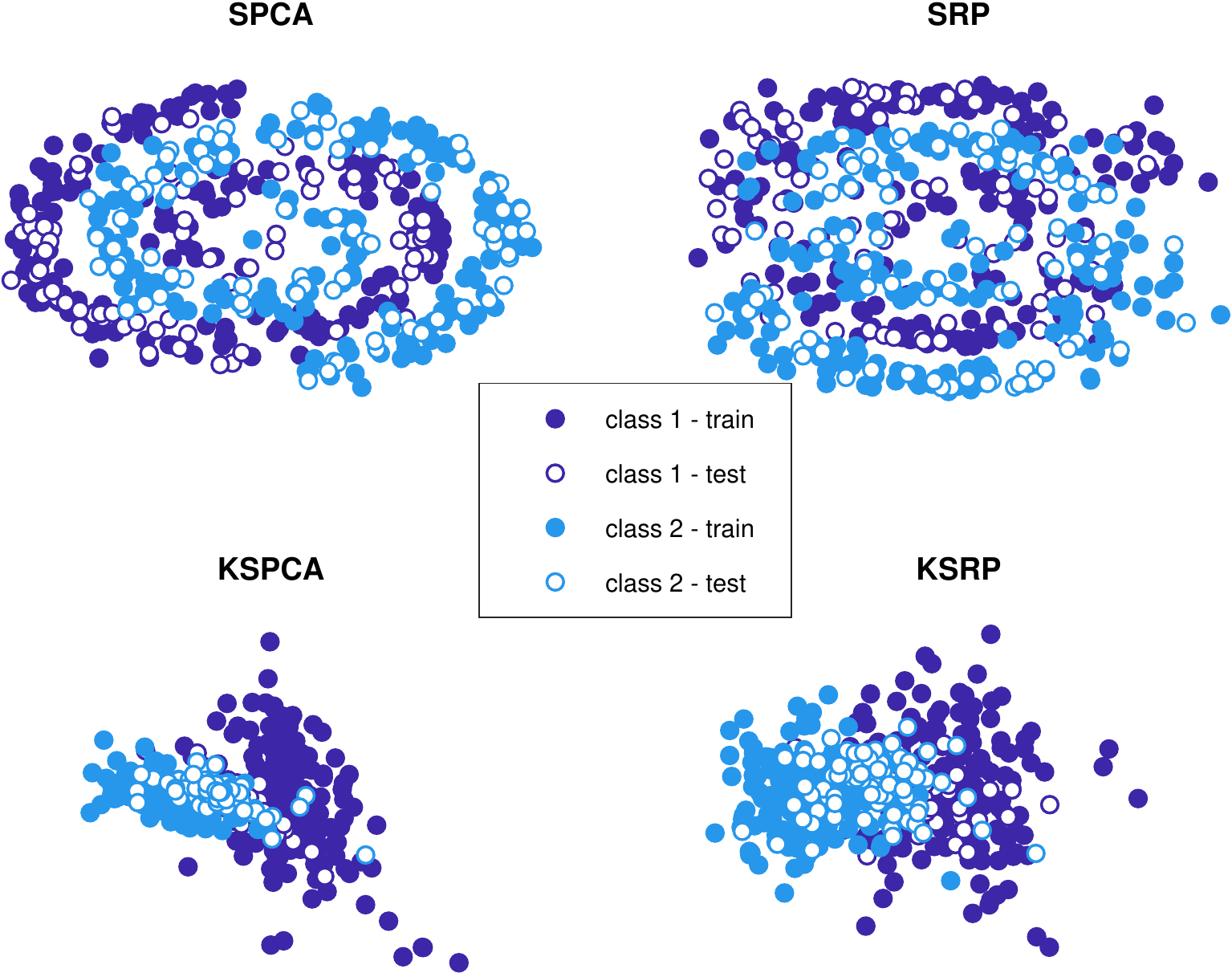}} \\ \\
      \multicolumn{2}{c}{\fbox{\includegraphics[height=0.36\textwidth, width=0.5\textwidth]{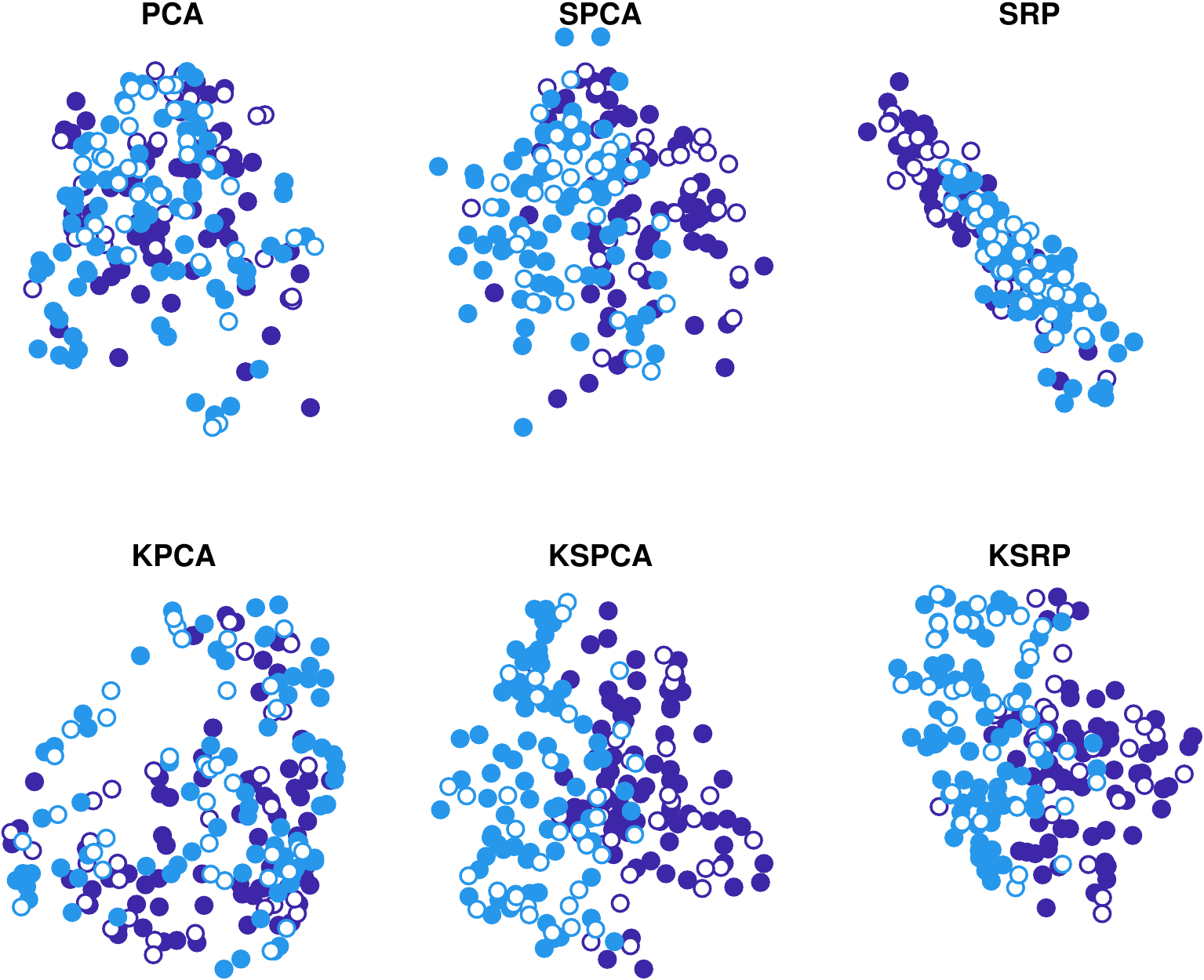}}}
  \end{array}$
  \caption{Visualization results for \textit{Binary XOR} (top-left), \textit{Spirals} (top-right), and \textit{UCI-Sonar} (bottom) datasets in $2$ dimensions.}
  \label{fig:visualization}
\end{figure*}

\subsection{Visualization}
\label{visualization}
First, the applicability of the proposed method on a data visualization task is examined. We compare performances on two synthetic datasets as well as a real-world dataset. The first synthetic dataset used was \textit{Binary XOR}, which comprised of $2$ classes distributed in fours clusters that are pair-wise positioned across from one another in $2$ dimensions. The second synthetic dataset was \textit{Spirals}, with each of the two arms corresponding to a separate class. For each synthetic datasets, we appended $8$ dimensions of random noise to all samples, which yielded $500$ samples in $10$-dimensions. Due to the relative positioning of the classes in the original space, these datasets are highly nonlinear and therefore we expect superior performance from KSPCA and KSRP compared to SPCA and SRP. The real-world data used here was \text{UCI-Sonar} from the UCI machine learning repository \cite{asuncion2007uci}, comprising $2$ classes with $208$ samples in $60$ dimensions. For all datasets we performed $70\%/30\%$ train/test split.

% The real-world data used here was the \text{USPS} handwritten digits dataset comprising $10$ digit classes. In order to not overpopulate our visualization plots, we randomly sampled $100/25$ training/testing samples from each classes, and each sample is $256$-dimensional.

Sample embeddings in $2$ dimensions are depicted in Figure \ref{fig:visualization}. Firstly, we observe that the embeddings for all datasets generalize to unseen test samples. For the highly nonlinear \textit{Binary XOR} and \textit{Spirals} datasets we can see that nonlinear approaches (i.e., KSPCA, KSRP) perform better by creating embeddings which congregates samples of the same class. Finally, for the \textit{UCI-Sonar} dataset we can see that a KSPCA and KSRP can embed a real-world dataset from $60$ dimensions to $2$ while keeping the classes well-separated. We additionally compare these embeddings with those obtained from label-agnostic PCA and KPCA, showing that the supervised approaches that leverage label information result in meaningful embeddings. In the next section, we present detailed time comparison for these methods, while quantitatively assessing embedding performance using 1-NN classification accuracy.

% For the USPS dataset, we observe similar positioning of classes relative to one another. In KSRP, however, the classes are more dispersed, agreeing with the derivation in Section \ref{supervised_random_projections} where we showed that KSRP embeddings are scaled versions of KSCPA according to the square-root of the singular values. While the embedding provided by KSPCA has fewer class overlaps and is potentially more favorable, KSRP is many times faster while still preserving a good outline of KSPCA. A potential future direction could explore a combination of these methods where a randomized KSRP approach is first employed to greatly reduce the dimensionality while preserving structure, and then KSPCA can be used to find tune the embeddings into neatly separated clusters per class.

\begin{figure*}[t!]
  $\begin{array}{rl}
      \fbox{\includegraphics[height=0.36\textwidth, width=0.45\textwidth]{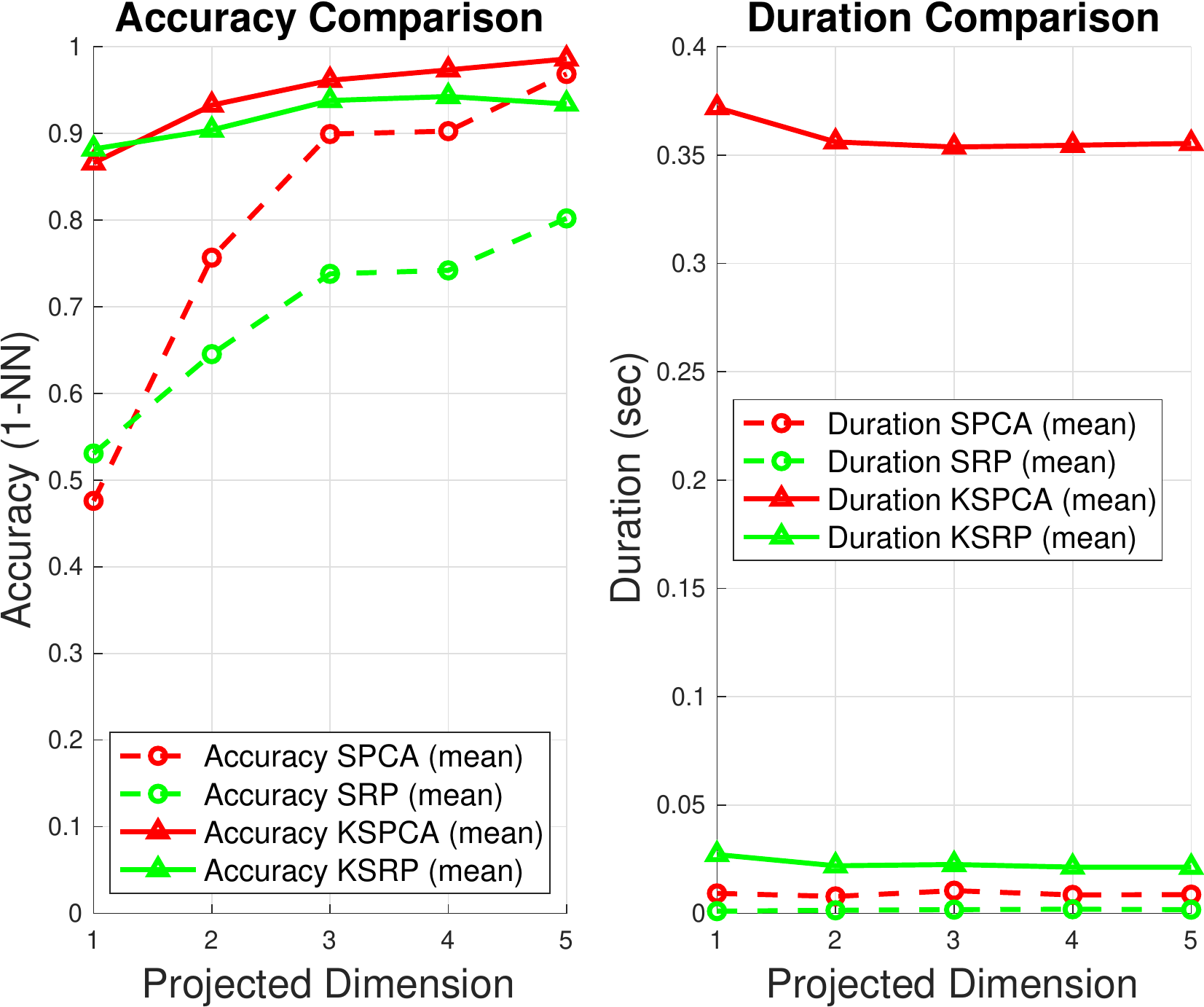}} &
      \fbox{\includegraphics[height=0.36\textwidth, width=0.45\textwidth]{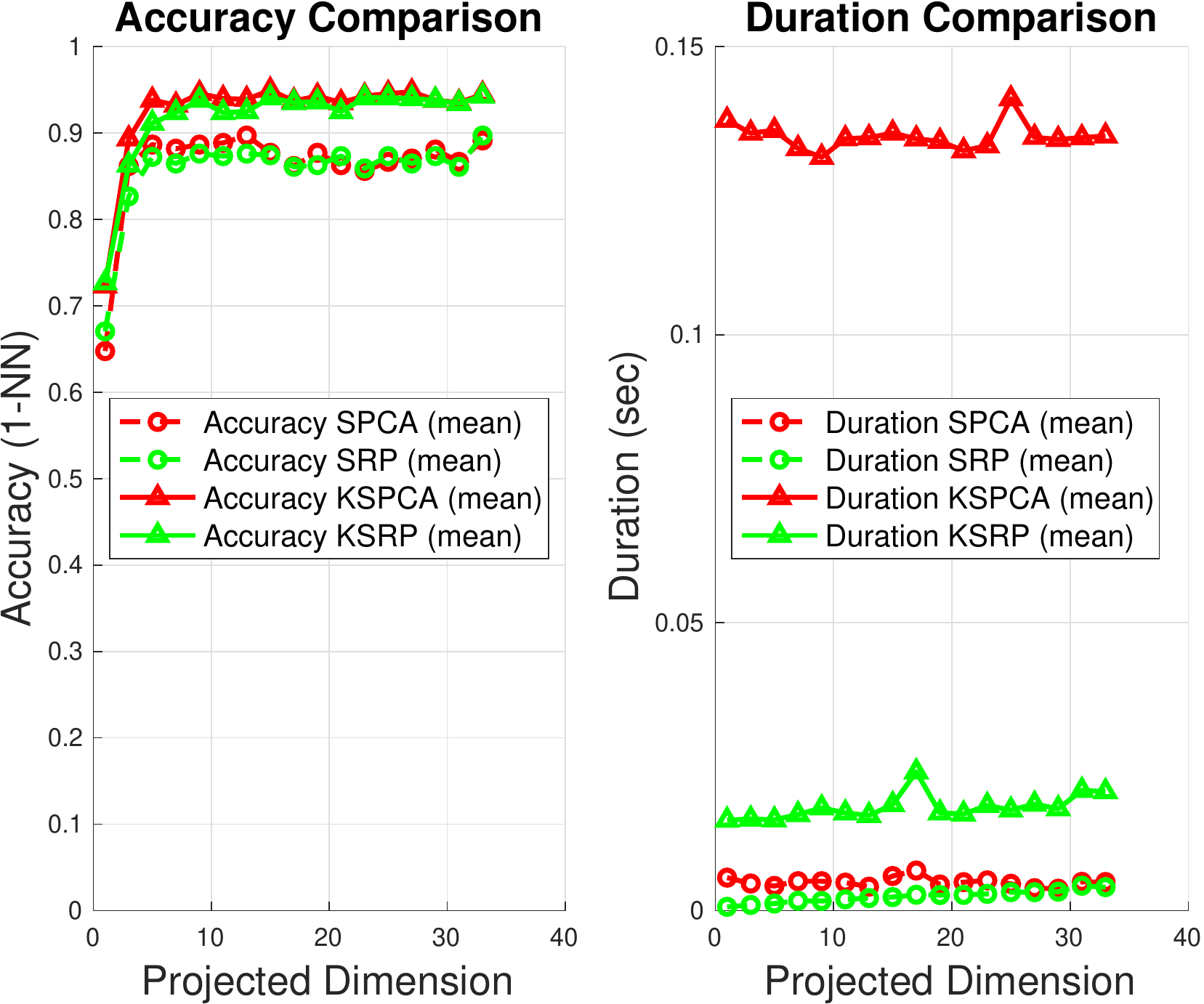}} \\ \\
      \multicolumn{2}{c}{\fbox{\includegraphics[height=0.36\textwidth, width=0.5\textwidth]{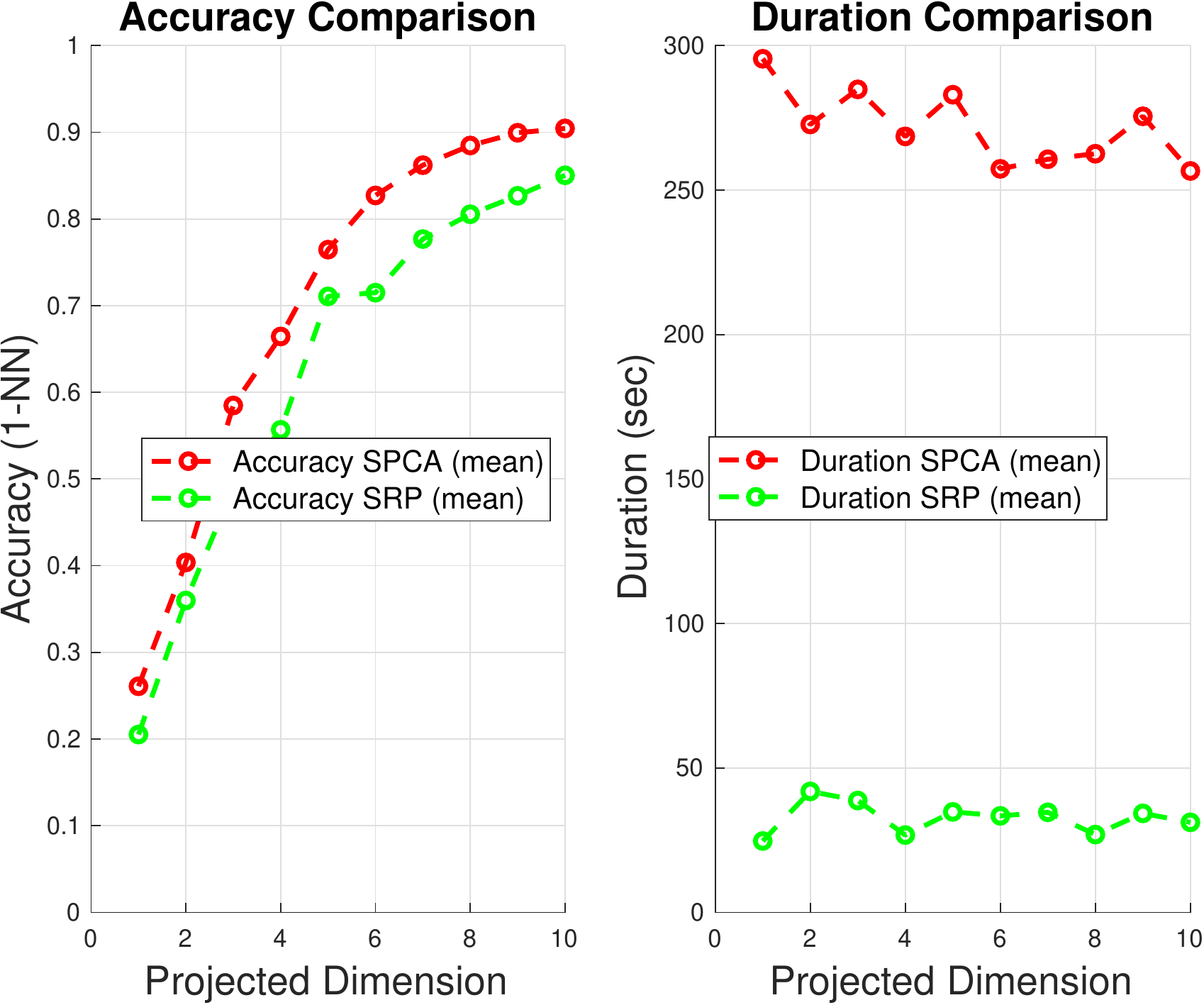}}}
  \end{array}$
  \caption{Classification results for \textit{Binary XOR} (top-left), \textit{UCI-Ionosphere} (top-right), and \textit{MNIST} (bottom) datasets in $2$ dimensions.}
  \label{fig:classification}
\end{figure*}

\subsection{Classification}

In this section we focus on classification problems and study the behavior of SRP and KSRP in comparison to SPCA and KSPCA. In \cite{barshan2011supervised}, SPCA and KSPCA were compared against other representative methods of supervised dimensionality reduction, and superior performance of these methods was demonstrated. Therefore, in this section, we suffice to only compare the proposed methods with SPCA and KSPCA. To do so, we run experiments on the synthetic \textit{Binary XOR} dataset described above, \textit{MNIST} \cite{MNIST} dataset, as well as \textit{UCI-Ionosphere} from the UCI machine learning repository \cite{asuncion2007uci}. The former contains $60,000$ training samples and $10,000$ testing samples in $784$ dimensions, and the latter contains $351$ samples in $34$ dimensions where we randomly perform a $70\%/30\%$ train/test split.

% We note a subtle restriction in choosing the projected space dimensionality, $k$. For a dataset with $c$ unique classes of object, assuming a delta kernel applied to the labels results in a labels kernel, $\zmat{L}$, with rank $c$. Furthermore, when $\zmat{L}$ is double centered (i.e., $\zmat{H} \zmat{L} \zmat{H}$), its rank drops to $c - 1$. From Section \ref{supervised_principal_component_analysis} we know that the rank of $\zmat{Q} = \zmat{X} \zmat{H} \zmat{L} \zmat{H} \zmat{X}^T$ essentially determines the maxmimum number of directions of variation in the data. Therefore, a decomposition of $\zmat{Q}$ resulting in the projection matrix $\zmat{U}$ will contain at most $c - 1$ meaningful eigenvectors; $\zmat{U}$ will contain $\min{d,n}$ eigenvectors of $\zmat{Q}$, but those corresponding to singular values equal $0$ carry no meaning. This suggests that while the original data may lie in a $d$-dimensional space, the projected dimension $k$ should be selected to be smaller than or equal to $c - 1$. This argument holds for SPCA and KSPCA, as well as for SRP and KSRP. One way to overcome the rank defficiency in $\zmat{L}$ is to add noise to the labels. We do this for \textit{UCI-Ionosphere} ($c = 2$) to measure performance on range of projected dimensions.

The results presented in Figure \ref{fig:classification} compare 1-NN and time performance on the three datasets, averaged over 30 runs. We make a number of interesting observations. Firstly, we note that increasing the number of random bases / projected dimension (i.e., $k$) results in better 1-NN performance. This is expected because with higher $k$s, we are retaining more information about the dataset in the embedding space. Assuming the test data is sampled from the same distribution as the training data (which may include the same noise distribution), higher $k$ should result in better performance 1-NN on the test set in the embedding space.

When comparing the time performances of various methods, we immediately notice the burdensome compute time required for KSPCA compared to that of KSRP and linear SPCA and SRP. We should keep in mind that, as mentioned in Section \ref{kernel_approximation}, there are kernel approximation schemes that are much faster than Random Kitchen Sinks which was used in our setup. This suggests that the efficiency gains observed here can be even more dramatic if we employ a kernel approximation method such as FastFood \cite{le2013fastfood}.

Bearing in mind the time complexities from Table \ref{table:time_complexities}, and considering that the number of samples in many datasets is typically larger than the dimensionality of the samples (i.e., $n > d$), this shows that while KSPCA has a total complexity of $\mathcal{O}(n^3)$, SPCA and SRP have a complexity of $\mathcal{O}(dn^2)$ and KSRP has a time complexity of $\mathcal{O}(\max\{k_x n^2,k_y n^2\})$. This aligns perfectly with the observed time duration results of Figure \ref{fig:classification}.

For the synthetic XOR plots (i.e., bottom row) of Figure \ref{fig:classification}, it agrees with our intuition that KSPCA and KSRP outperform their linear counterparts in lower dimensions. For \textit{MNIST}, we omit the results for KSPCA because we ran out of memory on a $64$GB machine when computing the data kernel matrix. We also noticed that linear approaches (SPCA, SRP) outperformed KSRP and therefore omitted these results in the figure for better illustration between SPCA and SRP. Across all plots, we see that randomized approaches have 1-NN performance very close to their exact counterpart (i.e., SPCA-SRP and KSPCA-KSRP pairs), while providing $5-20\times$ speed up in terms of wall-clock time.

\section{Conclusion}
\label{conclusion}

In this work, we propose a novel approach for efficiently finding an embedding of a large-scale dataset in the context of supervised dimensionality reduction. To achieve this, we were inspired by Supervised Principal Component Analysis (SPCA) and Kernel Supervised Principal Component Analysis (KSPCA) where an optimization function is solved in closed-form yielding an embedding that is maximally dependent on the labels. This work builds on the theory of SPCA and that of kernel approximation to construct Supervised Random Projections (SRP) and Kernel Supervised Random Projections (KSRP). We evaluated and compared these methods on five different datasets for visualization and classification, concluding that SRP and KSRP perform very competitively with SPCA and KSPCA: yielding a small drop in 1-NN classification performance while providing orders of magnitude better time performance.
 \newpage

\bibliographystyle{IEEEtran}
% \bibliographystyle{apalike}
% \bibliography{refs.bib}
\bibliography{refs.bib}

\end{document}